%% file: neurips_2026.tex
\documentclass{article}

\usepackage[dblblindworkshop,final]{neurips_2026}
\workshoptitle{Linguistic Principles for Foundation Models (LP4FM)}

\usepackage[utf8]{inputenc} 
\usepackage[T1]{fontenc}    
\usepackage{hyperref}       
\usepackage{url}            
\usepackage{booktabs}       
\usepackage{amsfonts}       
\usepackage{nicefrac}       
\usepackage{microtype}      
\usepackage{xcolor}         

\usepackage{amsmath}
\usepackage{latexsym}
\usepackage{graphicx}
\usepackage{inconsolata}
\usepackage{tikz}
\usetikzlibrary{shapes.geometric, arrows.meta, positioning, fit, calc, backgrounds}
\workshoptitle{Linguistic Principles for Foundation Models (LP4FM)}
\title{MITRA-Swarm: Injecting Explicit Linguistic Structure into Foundation Models for Classical Machine Translation}

\author{%
  Kayshav Bhardwaj \\
  Purdue University \\
  \texttt{bhardw43@purdue.edu} \\
  \And
  Pragun Seth \\
  UC Berkeley \\
  \texttt{pragunseth@berkeley.edu} \\
  \AND
  Kush Bhardwaj \\
  University of Illinois \\
  \texttt{kushb2@illinois.edu} \\
  \And
  Vijay Nagasamy \\
  UC Berkeley \\
  \texttt{vjnagasamy@gmail.com} \\
  \And
  Kurt Keutzer \\
  UC Berkeley \\
  \texttt{keutzer@berkeley.edu}
}

\begin{document}
\maketitle

\begin{abstract}
Classical Sanskrit poses three linguistic challenges monolithic foundation models do not resolve in one generation call: morphophonemic fusion (\textit{Sandhi}), dense nominal compounding (\textit{Samāsa}), and discourse-level zero-anaphora forced by poetic meter. We present \texttt{MITRA-Swarm}, a multi-agent architecture injecting explicit linguistic structure into an otherwise monolithic pipeline: a Sandhi-splitting agent, a parallel array of generalist (\texttt{gemma-4-26B-A4B-it}) and domain-specialized (\texttt{gemma-2-mitra-it}) translators, and a discourse-aware Ellipsis Restorer that consults corpus-derived register tags to avoid inventing active-voice subjects in impersonal legal and philosophical registers. Quality is estimated by two strictly separated judges: a routing judge ($J_{\text{route}}$, same model family as the generators) driving an internal critique loop, and an independent reporting judge ($J_{\text{report}}=$\texttt{gpt-4o-mini}, no shared vendor lineage) scoring every output exactly once. On a stratified benchmark of 1,000 verses across five registers (23,990 scored verdicts, 12-condition pre-registered ablation), the full swarm reaches $85.90$ MQM against a monolithic baseline of $85.27$ ($\Delta=+0.63$, $p_{\text{adj}}=0.0018$, 95\% CI $[+0.30,+0.97]$), and isolated discourse restoration alone accounts for $+0.44$ MQM ($p_{\text{adj}}=0.0288$), outperforming a call-count-matched self-refinement baseline by $+0.51$ MQM ($p_{\text{adj}}=0.0288$). Severe per-verse failures (MQM~$<80$) fall from $9.0\%$ to $7.6\%$, at roughly $9\times$ the per-verse inference cost of the baseline. Beyond these headline effects, we report two findings about evaluation itself: $J_{\text{route}}$ scores its own family's outputs $2$--$7$ MQM higher than $J_{\text{report}}$ on every one of the 12 conditions, empirically motivating the two-judge split; and a preflight screen of a 192{,}151-entry retrieval index against all 12 evaluation works verifies 0 reference leaks across 269 runtime checks. We report these as small effects on a metric that saturates near 85-86 across every condition.
\end{abstract}

\section{Introduction}

Sanskrit is the primary linguistic repository for a vast corpus of classical Indian philosophy, science, and literature, yet remains severely under-resourced in modern NLP: the lack of large parallel corpora and the language's grammatical complexity have historically limited MT quality \citep{haddow-etal-2022-survey, nllb2022}. Foundation models trained overwhelmingly on high-resource, concatenative languages do not, by default, carry the inductive biases classical Sanskrit requires, i.e. no mechanism for undoing phonetic fusion at word boundaries, no structured representation of deeply nested compounds, and no discourse memory built to restore an argument elided clauses earlier for metrical reasons, which in turn produces translations marred by omission, semantic drift, and hallucination \citep{ji2023survey}.

Three linguistic properties make this failure systematic rather than incidental:
\begin{itemize}
    \item \textbf{Sandhi (Morphophonemic Fusion):} phonetic mergers at word boundaries obscure lexical boundaries without pre-translation segmentation, causing out-of-vocabulary and dictionary-lookup failures.
    \item \textbf{Samāsa (Nominal Compounding):} compounds, sometimes spanning entire clauses, leave the syntactic and semantic relations between their constituents implicit.
    \item \textbf{High Ellipsis (Zero Anaphora):} metrical verse systematically omits subjects, objects, even verbs, requiring inference from the surrounding discourse.
\end{itemize}
A monolithic model attempts segmentation, reordering, lexical translation, and discourse-level argument restoration in a single generation call; without explicit modular guidance it loses local morphosemantic structure, hallucinates relations between compound members, and invents target-language arguments rather than interrogating the source. \texttt{MITRA-Swarm} instead injects each linguistic principle as an explicit, separately-verifiable pipeline stage:
\begin{enumerate}
    \item \textbf{A morphological Sandhi Parser} decomposes phonetic mergers and compounds into a structured morphosemantic guide before translation.
    \item \textbf{A parallel generator array} of a generalist frontier model (\texttt{gemma-4-26B-A4B-it}) and a domain-specialized fine-tune (\texttt{gemma-2-mitra-it}) compares specialization against raw scale rather than assuming it.
    \item \textbf{A discourse-level Ellipsis Restorer with register-aware voice guarding} (Section~\ref{sec:registers}) restores zero-anaphora referents in narrative registers while suppressing invented active-voice subjects in impersonal legal (\textit{Dharmaśāstra}) and philosophical (\textit{Sūtra}) injunctions.
    \item \textbf{A two-judge quality architecture}: a routing judge $J_{\text{route}}$ (shares a model family with the generators) drives an internal critique loop; an independent reporting judge $J_{\text{report}}$ (shares no vendor lineage with any generator or $J_{\text{route}}$) scores every finished output exactly once, off the critical path, for the numbers in this paper.
\end{enumerate}
When $J_{\text{route}}$'s estimate falls below threshold, a Critique Agent diagnoses errors, a Tool Selector queries whatever retrieval tools are genuinely backed by data, and a Refinement Agent rewrites the translation.

Our contributions:
\begin{itemize}
    \item \textbf{Architecture.} Agents that each target one linguistic failure mode of monolithic models: fusion, compounding, and discourse ellipsis, rather than leaving all three to a single unguided generation call.
    \item \textbf{Empirical scale and significance.} A pre-registered 12-condition ablation on $N=1{,}000$ verses (23,990 scored verdicts): full swarm beats the monolithic baseline by $+0.63$ MQM ($p_{\text{adj}}=0.0018$) and a call-count-matched self-refinement baseline by $+0.51$ MQM ($p_{\text{adj}}=0.0288$), via paired approximate randomization with Holm--Bonferroni correction and a measured noise floor rather than an uncorrected bootstrap-tail probability.
    \item \textbf{An isolated discourse-syntax finding.} Ablating only the Ellipsis Restorer costs $-0.44$ MQM ($p_{\text{adj}}=0.0288$), representing the single largest individually-significant component effect, and the clearest evidence that a linguistically-motivated discourse module, not scale or generic self-critique, is doing the work.
    \item \textbf{Evaluation and decontamination rigor.} We directly quantify LLM-judge self-preference ($J_{\text{route}}$ scores its own family's outputs $2$–$7$ MQM higher than $J_{\text{report}}$ on every condition, which is why only $J_{\text{report}}$ is reported) and screen a 192{,}151-entry retrieval index for reference leakage (0 leaks / 269 runtime checks).
\end{itemize}

\section{Related Work}

\paragraph{Sanskrit Computational Linguistics}
Prior work builds on foundational lexical-database and morphological-analysis resources \citep{huet2004design}, and more recently on neural sequence-to-sequence splitters \citep{dave2021neural} and transformer tokenizers such as TransLIST \citep{sandhan-etal-2022-translist}, integrated into tools like SanskritShala \citep{sandhan-etal-2023-sanskritshala} and corpora such as the Digital Corpus of Sanskrit \citep{dcs}.

\paragraph{LLMs for Machine Translation}
General-purpose LLMs excel on high-resource languages \citep{gpt-mt-2023} but struggle where parallel data is scarce \citep{haddow-etal-2022-survey}. Two-stage fine-tuning, monolingual pretraining followed by a small amount of high-quality parallel data (ALMA, \citealp{xu2024alma}), motivates our domain-aligned \texttt{gemma-2-mitra-it} backbone \citep{mitra-model-paper}; scaling efforts like NLLB-200 \citep{nllb2022} expand coverage without addressing classical-language-specific needs like zero-anaphora resolution \citep{wang-etal-2023-survey}.

\paragraph{Multi-Agent Systems and Self-Refinement}
Multi-agent debate among independent LLM instances improves factuality and reasoning \citep{du2023improving}; TransAgents \citep{wu-etal-2025-perhaps} showed multi-role literary translation with specialized agents preferred by humans despite lower n-gram scores. Self-refinement \citep{madaan2023selfrefine} is directly implemented, call-count-matched, as our \texttt{self\_refine} ablation. Tool-augmented reasoning \citep{yao2023react, schick2023toolformer} and retrieval augmentation \citep{khandelwal2021nearest} motivate Section~\ref{sec:tools}, where we report what happens when tool-use is restricted to tools actually backed by data.

\paragraph{Reference-Free Evaluation and Judge Reliability}
Reference-based metrics have documented validity limits \citep{reiter-2018-structured}; reference-free QE \citep{rei-etal-2022-cometkiwi} and fine-grained error-span detection \citep{guerreiro-etal-2024-xcomet} address this, as do LLM-as-judge frameworks \citep{kocmi-federmann-2023-large, kocmi-federmann-2023-gemba} aided by chain-of-thought prompting \citep{wei2022chainofthought}. Such judges also exhibit documented biases like self-preference \citep{zheng2023judging}: a judge sharing a model family with the system it evaluates is one instance, and Section~\ref{sec:judge-validation} measures it directly rather than assuming it away.

\section{Methodology}

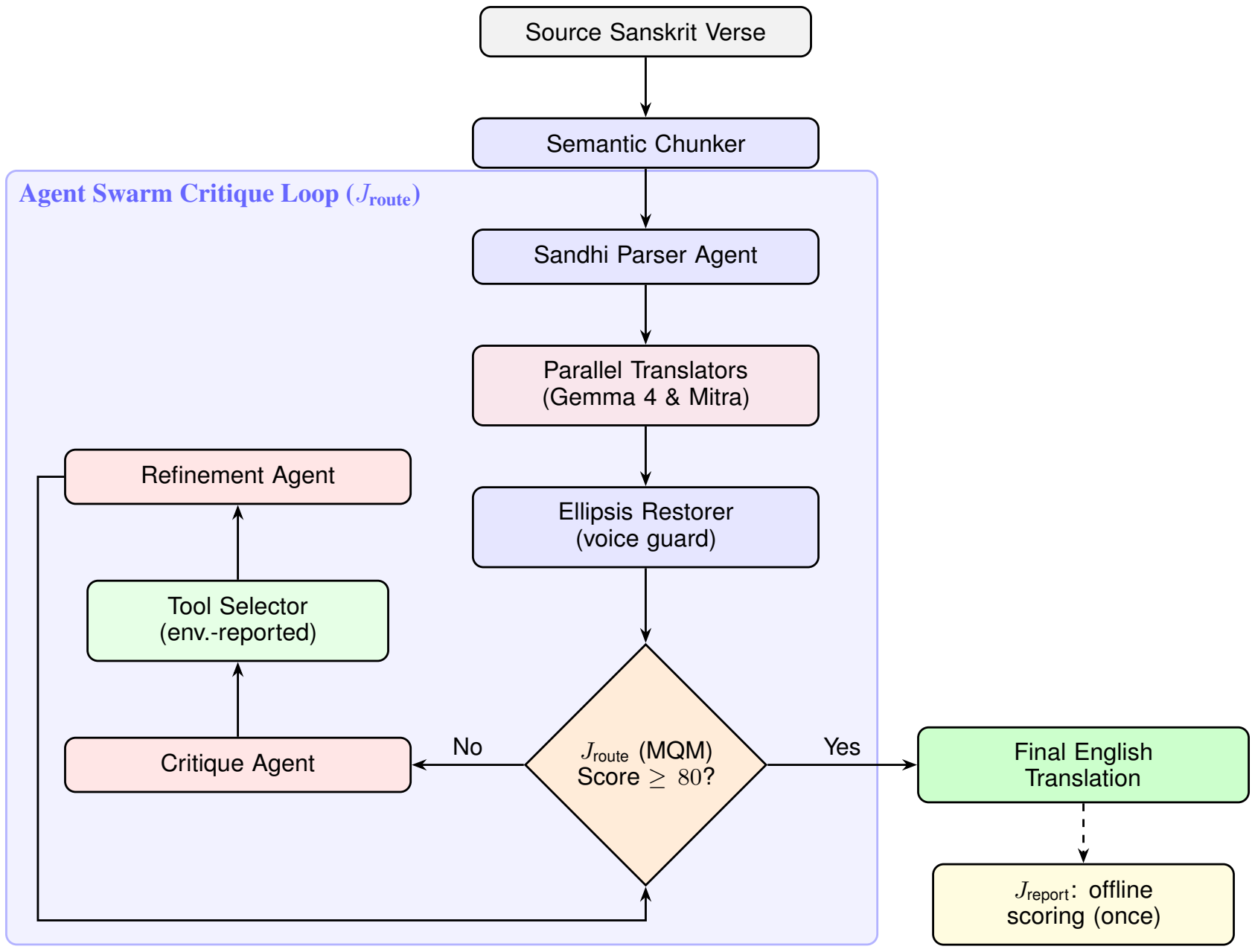
\begin{figure}[t]
\centering
\resizebox{0.8\linewidth}{!}{
\begin{tikzpicture}[
    node distance=1.2cm and 0.9cm,
    font=\small\sffamily,
    base/.style = {rectangle, rounded corners, draw=black, thick, text centered, inner sep=6pt},
    agent/.style = {base, fill=blue!10, text width=4.2cm},
    data/.style = {base, fill=gray!10, text width=4cm},
    decision/.style = {diamond, draw=black, thick, fill=orange!15, text centered, text width=2.6cm, inner sep=0pt},
    tool/.style = {base, fill=green!10, text width=3.6cm},
    arrow/.style = {thick, -{Stealth[length=2mm]}},
    dashed_arrow/.style = {thick, dashed, -{Stealth[length=2mm]}}
]

\node (input) [data] {Source Sanskrit Verse};
\node (chunk) [agent, below=0.8cm of input] {Semantic Chunker};
\node (sandhi) [agent, below=0.8cm of chunk] {Sandhi Parser Agent};
\node (drafting) [agent, fill=purple!10, below=0.8cm of sandhi] {Parallel Translators (Gemma 4 \& Mitra)};
\node (ellipsis) [agent, below=0.8cm of drafting] {Ellipsis Restorer (voice guard)};
\node (judge) [decision, below=1cm of ellipsis] {$J_{\text{route}}$ (MQM)\\Score $\ge 80$?};

\node (output) [data, fill=green!20, right=2cm of judge] {Final English\\Translation};
\node (report) [tool, fill=yellow!15, below=0.8cm of output] {$J_{\text{report}}$: offline scoring (once)};

\node (critique) [agent, fill=red!10, left=1.5cm of judge] {Critique Agent};
\node (tools) [tool, above=1cm of critique] {Tool Selector (env.-reported)};
\node (refinement) [agent, fill=red!10, above=1cm of tools] {Refinement Agent};

\draw [arrow] (input) -- (chunk);
\draw [arrow] (chunk) -- (sandhi);
\draw [arrow] (sandhi) -- (drafting);
\draw [arrow] (drafting) -- (ellipsis);
\draw [arrow] (ellipsis) -- (judge);

\draw [arrow] (judge) -- node[anchor=south] {Yes} (output);
\draw [arrow] (judge) -- node[anchor=south] {No} (critique);
\draw [dashed_arrow] (output) -- (report);

\draw [arrow] (critique) -- (tools);
\draw [arrow] (tools) -- (refinement);

\draw [arrow] (refinement.west) -- ++(-0.35,0) |- ([yshift=-0.45cm]judge.south) -- (judge.south);

\begin{scope}[on background layer]
    \node [draw=blue!30, fill=blue!5, thick, rounded corners, fit=(sandhi) (drafting) (ellipsis) (critique) (tools) (refinement) (judge), inner sep=22pt] (swarm) {};
    \node [anchor=north west, font=\bfseries\color{blue!60}, inner sep=5pt] at (swarm.north west) {Agent Swarm Critique Loop ($J_{\text{route}}$)};
\end{scope}

\end{tikzpicture}
}
\caption{System architecture of the \texttt{MITRA-Swarm} pipeline. $J_{\text{route}}$ (same model family as the generators) drives the internal critique loop; $J_{\text{report}}$ (independent vendor, \texttt{gpt-4o-mini}) scores the finished output exactly once, off the critical path, and is the only judge whose scores are reported as results (Section~\ref{sec:judge-validation}).}
\label{fig:pipeline}
\end{figure}

The \texttt{MITRA-Swarm} pipeline decomposes translation into chunking, morphosemantic preprocessing, parallel drafting, discourse-aware argument restoration, automated quality routing, and agentic refinement, illustrated in Figure~\ref{fig:pipeline} and detailed with exact prompts in Appendix~\ref{sec:appendix}.

\subsection{Document Segmentation (Chunking)}
To manage context-window limits and preserve local discourse cohesion \citep{liu-etal-2024-lost}, the source document is segmented at classical sentence/verse terminators (\textit{daṇḍa} \texttt{|}, \textit{dvipada daṇḍa} \texttt{||}), a sliding token window ($W=200$, overlap $O=40$) for unpunctuated prose, or an LLM-driven semantic chunker for narrative boundaries.

\subsection{Morphological Pre-processing (Sandhi Parser)}
A Sandhi Parser Agent decomposes phonetic mergers and compounds (\textit{Samāsa}; \citealp{dave2021neural, sandhan-etal-2022-translist}) into a structured morphosemantic guide, threaded through a typed \texttt{TranslationRequest} into every downstream translator call. \texttt{gemma-4-26B-A4B-it} accepts the guide via its chat template; \texttt{gemma-2-mitra-it} uses a legacy completion-style prompt for which guide injection was evaluated empirically and disabled, so the \texttt{guide\_only}/\texttt{guide\_ellipsis} contrasts reflect that per-model choice rather than a uniform assumption.

\subsection{Parallel Translation Array}
The chunk is translated concurrently by a parallel array of models,
\begin{equation}
    \mathbf{T} = \{t_{\text{specialized}}, t_{\text{frontier}}, \dots\},
\end{equation}
comprising the domain-specialized \texttt{gemma-2-mitra-it} \citep{mitra-model-paper} and the generalist frontier \texttt{gemma-4-26B-A4B-it} \citep{gemma4google2026}, both served locally via vLLM. Each model generates a draft translation independently, and the routing judge selects among candidates.

\subsection{Ellipsis and Zero-Anaphora Argument Restoration with Register-Aware Voice Guarding} \label{sec:registers}
The Ellipsis Restorer operates over a sliding discourse window of the preceding/succeeding chunks, $\mathbf{C}_{\text{discourse}} = \{c_{i-k}, \dots, c_{i}\}$, identifying zero-anaphora and interpolating the implied argument. Naively restoring an elided subject risks inventing a third-person actor in passages that are impersonal or injunctive by design, where restoration could convert a \textit{Dharmaśāstra} injunction into an active clause with a hallucinated subject. \texttt{MITRA-Swarm} mitigates this with a \textbf{voice-preservation guard}: every source work ID is mapped, via a corpus-derived, manually verified register table (\texttt{registers.json}; six registers (\textit{vedic}, \textit{epic}, \textit{legal}, \textit{sutra}, \textit{kavya}, \textit{shastra}), covering over 100\% of the corpus's 52 work IDs, rather than inferred from a keyword heuristic), to a \texttt{voice\_preserving} flag. \textit{Legal} and \textit{sutra} are flagged voice-preserving as characteristically impersonal/passive by genre; the Ellipsis Restorer receives the literal genre hint \texttt{"legal"} for any work so flagged, and the raw register name otherwise, which measurably changes its restoration behavior (Section~\ref{sec:qualitative}) without a separate model or prompt per register.

\subsection{Two-Judge Quality Architecture} \label{sec:two-judge}
Quality estimation is split into two roles, enforced at startup by an automated model-family check. \textbf{$J_{\text{route}}$ (Routing Judge):} a locally-hosted \texttt{gemma-4-26B-A4B-it} instance, sharing a model family with the generators, scores intermediate drafts $S_j \in [0,100]$ against threshold $T_{\text{judge}}=80$ to decide whether the critique loop fires; because it shares a family with what it evaluates and is queried repeatedly during generation, its scores are an internal control signal only, never reported as a result. \textbf{$J_{\text{report}}$ (Reporting Judge):} an independent offline judge (\texttt{openai:gpt-4o-mini}), sharing no vendor lineage with either generator or $J_{\text{route}}$, invoked exactly once per finished output on a separate machine from the generation cluster. Every number in Section~\ref{sec:results} is a $J_{\text{report}}$ score.

Both judges use a GEMBA-style rubric returning structured error spans (category, severity, verbatim span, explanation) rather than a bare number. The headline MQM is \emph{recomputed} deterministically from that list using fixed severity weights (critical $=-25$, major $=-5$, minor $=-1$, floored at 0), not the model's own top-line arithmetic, which is retained separately for the judge-validation diagnostic (Section~\ref{sec:judge-validation}). A failed judge call yields a missing score, excluded from all aggregates, rather than a sentinel $0.0$ that would otherwise depress the mean and minimum and spuriously trigger the critique loop.

\subsection{Agentic Feedback and Tool Retrieval} \label{sec:tools}
If $S_j < T_{\text{judge}}$, a Critique Agent diagnoses errors, a Tool Selector queries whichever tools are genuinely backed by data, and a Refinement Agent synthesizes the draft, critique, and retrieved context into a revised translation, re-scored up to $M$ iterations. The inventory is reported \emph{as offered}: a tool with no backing index is absent rather than present and silently returning a canned string. On the air-gapped GH200 cluster used here, that inventory reduces to \textbf{\texttt{TRANSLATION\_MEMORY}}, which performs fuzzy character-6-gram matching against a decontaminated corpus index (Section~\ref{sec:decon}). A \texttt{DICTIONARY} tool (offline lexicon lookup, with entries OCR-sanitized to strip 19th-century scanning artifacts before reaching the Refinement Agent) exists but was not backed by a snapshot in this run; \texttt{ALIGNMENT} and live \texttt{SEARCH} require, respectively, an alignment index and network access unavailable offline. A Dedup Agent stitches overlapping segments pairwise, $t_{\text{merged}} = \text{Stitch}(t_{i-1}, t_i)$, feeding only the tail of the running translation (last $L=600$ characters) so per-step input stays bounded rather than growing with the number of merged segments.

\section{Experimental Setup}

\subsection{Evaluation Dataset and Stratification} \label{sec:dataset}
Experiments are conducted on a stratified sample of $N=1{,}000$ verses drawn, proportionally by register, from the 5{,}552-verse test partition (12 works) of the \texttt{MitraSamgraha} corpus \citep{nehrdich2026mitrasamgrahacomprehensiveclassicalsanskrit}. Table~\ref{tab:registers} reports the realised sample against the full test-partition population for each of the five registers represented in this partition; the register-to-work-ID mapping is manually verified (100\% work-ID coverage, 0 unmapped) rather than inferred from a keyword heuristic.

\begin{table}[h]
\centering
\small
\caption{Register composition of the $N=1{,}000$ evaluation sample, proportional to its population in the 5{,}552-verse test partition. \textit{Voice-preserving} marks registers routed to the impersonal-voice guard in the Ellipsis Restorer (Section~\ref{sec:registers}).}
\begin{tabular}{lp{3.3cm}rrc}
\toprule
Register & Description & Pop. & Sample & Voice-preserv.\\
\midrule
\texttt{epic} & Itihāsa/Purāṇa narrative & 1{,}720 & 310 & -- \\
\texttt{sutra} & Philosophical sūtra, commentary & 1{,}488 & 268 & \checkmark \\
\texttt{vedic} & Vedic saṃhitā/upaniṣad; archaic & 1{,}000 & 180 & -- \\
\texttt{kavya} & Ornate metrical poetry & 844 & 152 & -- \\
\texttt{legal} & Dharmaśāstra, nibandha & 500 & 90 & \checkmark \\
\midrule
Total & & 5{,}552 & 1{,}000 & \\
\bottomrule
\end{tabular}
\label{tab:registers}
\end{table}

This sample size balances statistical power with the cost of a 12-condition ablation (12,000 full pipeline runs). Because the corpus is drawn from 12 works with substantial internal stylistic variation, register-level results are an average over that variation, not a claim of uniformity within it.

\subsection{Data Decontamination and Preflight Reference-Echo Screening} \label{sec:decon}
To ensure rigorous evaluation and prevent data leakage, the \texttt{TRANSLATION\_MEMORY} index contains $192{,}151$ segment-level entries, with all 12 evaluation work IDs excluded at build time ($125{,}665$ entries excluded; $0$ evaluation work IDs reachable in the live index). A runtime \textbf{reference-echo detector} computes character 6-gram F-score similarity between every retrieved string and the gold reference that triggered it, flagging anything $\ge0.6$; across a 200-item preflight check and 269 runtime retrievals, the maximum observed similarity is $0.375$, with $0$ echoes flagged. We separately audit translator outputs against the gold reference for suspected training-data memorization (independent of retrieval): $0.21\%$ of \texttt{gemma} and $0.70\%$ of \texttt{mitra} outputs exceed a $0.6$ "suspect" threshold, with $0.15\%$ of \texttt{mitra} outputs exceeding a stricter $0.8$ threshold; inspection shows short sentences (e.g., \textit{"He makes offering(s) to four deities"}) with only a one-word paraphrase gap, not retrieval-time leakage.

\subsection{Models and Hardware}
The generator array comprises \texttt{gemma-4-26B-A4B-it} (26B-parameter MoE, 4B active) and \texttt{gemma-2-mitra-it}, both served locally via vLLM on NVIDIA GH200 nodes; $J_{\text{route}}$ runs on the same endpoint. $J_{\text{report}}$ runs offline, after generation, against the OpenAI API (\texttt{gpt-4o-mini}), since the generation cluster is air-gapped.

\subsection{Hyperparameters and Compute Budget}
Translation agents decode at temperature $0.0$; critique/judge/tool-selector agents at $0.1$; serving context window is capped at $4{,}096$ tokens (with a $2{,}048$-token output cap) to optimize batch throughput across verse segments, well within the models' native context windows. Table~\ref{tab:compute} compares per-verse token budget and latency for the three conditions central to the main results (full per-condition table in Appendix Table~\ref{tab:compute-full}), measured directly from run instrumentation, not estimated. Across the full run, generation consumed $17.61$M tokens ($10.71$M prompt, $6.90$M completion) across $85{,}509$ LLM calls at a $54.9\%$ cache hit rate; the routing judge alone accounts for $17{,}713$ calls ($12.57$M tokens), since $J_{\text{route}}$ is queried once per candidate per stage rather than once per verse.

\begin{table}[h]
\centering
\small
\caption{Per-verse compute, key conditions (measured, not estimated). Full swarm costs $9.2\times$ baseline tokens for $+0.63$ MQM.}
\begin{tabular}{lrrrr}
\toprule
Configuration & Prompt tok. & Compl. tok. & Total/verse & Latency (s) \\
\midrule
baseline & 91 & 49 & 140 & 2.5 \\
self\_refine & 164 & 89 & 253 & 4.9 \\
bo\_n & 455 & 258 & 713 & 12.8 \\
full\_swarm & 828 & 462 & 1290 & 7.1 \\
\bottomrule
\end{tabular}
\label{tab:compute}
\end{table}

\subsection{Ablation Matrix} \label{sec:ablation_configs}
We pre-register a 12-condition ablation matrix (full active-component flags in Appendix Table~\ref{tab:configmatrix}), isolating each architectural component against a monolithic baseline, a paired test-retest control, a placebo re-judging control, and two compute-matched controls (best-of-$N$ and generic self-refinement): \texttt{baseline} (monolithic single-pass, no guide/ellipsis/loop); \texttt{baseline\_dup} (different seed; the noise floor); \texttt{placebo\_rejudge} (baseline re-judged; judge-side noise only); \texttt{bo\_n} (compute-matched best-of-5, temperature 0.7); \texttt{self\_refine} (call-count-matched generic self-critique, no role specialization); \texttt{guide\_only} (Sandhi guide alone); \texttt{guide\_ellipsis} (guide + Ellipsis Restorer, no loop/tools); \texttt{no\_guide}, \texttt{no\_ellipsis} (the isolated discourse-restoration ablation), \texttt{no\_critique}, \texttt{no\_tools} (full swarm minus one component each); and \texttt{full\_swarm}, the complete architecture (Section~\ref{sec:registers}--\ref{sec:tools}).

\subsection{Metrics and Statistical Plan} \label{sec:stats}
Quality is \textbf{MQM}, a reference-free $[0,100]$ score from $J_{\text{report}}$ (Section~\ref{sec:two-judge}). Significance uses a \textbf{paired approximate randomization test} \citep[the standard for MT system comparison rather than a bootstrap-tail probability;][]{riezler-maxwell-2005} at $R=10{,}000$, with 95\% BCa bootstrap CIs ($B=10{,}000$); all 10 pre-registered contrasts are corrected jointly with \textbf{Holm--Bonferroni}. Because mean MQM is compressed near 85--86 for every condition, we also report a \textbf{measured noise floor}: the mean absolute difference between \texttt{baseline} and \texttt{baseline\_dup} is $0.21$ MQM, and a contrast is \emph{reportable} only if it is both significant \emph{and} exceeds this floor; a significant result smaller than the chance disagreement between two identical systems is a measurement of the harness, not the method.

\section{Results and Analysis} \label{sec:results}

\subsection{Main Ablation Results}
Table~\ref{tab:ablation} reports the full 12-condition matrix; Table~\ref{tab:contrasts} reports the 10 pre-registered contrasts derived from it.

\begin{table}[t]
\centering
\scriptsize
\caption{Ablation results on the $N=1{,}000$ stratified \texttt{MitraSamgraha} benchmark. MQM from $J_{\text{report}}$ (\texttt{gpt-4o-mini}); P10/P25 are the 10th/25th percentiles; \%$<80$ is the severe-failure rate.}
\input{paper/tables/ablation_main.tex}
\label{tab:ablation}
\end{table}

\begin{table}[t]
\centering
\scriptsize
\caption{Pre-registered contrasts, paired approximate randomization test, $R=10{,}000$, Holm--Bonferroni adjusted over the family of 10. $^{*}$ significant and exceeds the $0.21$ MQM noise floor. $^{\dagger}$ below the noise floor (not distinguishable from run-to-run variation regardless of nominal $p$-value).}
\input{paper/tables/contrasts.tex}
\label{tab:contrasts}
\end{table}

Four trends are reportable under the criterion in Section~\ref{sec:stats} (significant and above the noise floor). \textbf{(1)} The full swarm beats the monolithic baseline by $+0.63$ MQM ($p_{\text{adj}}=0.0018$, CI $[+0.30,+0.97]$, $d_z=0.12$, 397/294/308 W/L/T), which is real but small in absolute terms (both means sit at 85.3--85.9 of 100), which we report plainly. \textbf{(2)} Removing only the Ellipsis Restorer costs $-0.44$ MQM relative to the full swarm ($p_{\text{adj}}=0.0288$, $[+0.12,+0.73]$, 323/232/444 W/L/T), constituting the single largest individually-significant component effect, confirming that discourse restoration operates effectively in isolation. \textbf{(3)} The swarm beats a call-count-matched generic self-refinement baseline by $+0.51$ MQM ($p_{\text{adj}}=0.0288$, 389/301/309 W/L/T), attributable to role specialization and discourse/morphology structure rather than extra inference calls alone. \textbf{(4)} Adding the Ellipsis Restorer on top of the Sandhi guide alone yields $+0.69$ MQM ($p_{\text{adj}}=0.001$, $[+0.39,+1.00]$, $d_z=0.14$), the largest effect size in the matrix, consistent with (2).

The other six contrasts are \emph{not} reportable, and a null from a well-controlled comparison is itself informative: the Sandhi guide alone, against the full swarm ($-0.01$) or the baseline ($-0.04$, \texttt{guide\_only}); the critique loop in isolation ($-0.02$); tool retrieval ($+0.01$; Section~\ref{sec:tools} explains why, given the tool inventory actually available); the placebo re-judging control ($-0.01$, confirming judge stability under repeat scoring); and full swarm vs.\ best-of-$N$ ($+0.32$, $p_{\text{adj}}=0.460$, numerically positive and above the floor but not significant, though a large enough compute-matched sampling budget may close some of this gap).

\textbf{These findings replicate under a fully deterministic, reference-based metric.} To check they do not depend on a single judge's preferences, we additionally compute chrF++ \citep{popovic-2017-chrf} via \texttt{sacrebleu} (word order 2) over the same selected outputs. The full swarm's advantage over baseline holds under this judge-independent metric ($29.27\to30.33$ chrF++, $p_{\text{adj}}=0.0010$), as does isolated ellipsis restoration ($+0.73$, $p_{\text{adj}}=0.0010$) and the gain over self-refine ($+0.78$, $p_{\text{adj}}=0.0010$); the same noise-floor check that gates the MQM contrasts flags marginal seed-to-seed chrF variance on its own ($p=0.033$), smaller than the reported effects but disclosed rather than omitted.

Restricting to baseline, the isolated ellipsis ablation, and the full swarm (full 12-condition breakdown in Appendix Table~\ref{tab:registers-full}), the full-swarm gain over baseline is largest in \textit{vedic} ($83.6\to84.5$, $+0.9$; heaviest ellipsis, most archaic morphology), moderate in \textit{kavya} ($+0.4$), \textit{legal}, and \textit{sutra} ($+0.3$ each), and flat in \textit{epic} ($+0.0$). This pattern is consistent with, though not definitive proof of, the discourse-restoration mechanism in Section~\ref{sec:registers}: the voice-preserving registers (\textit{legal}, \textit{sutra}) do not show the largest numeric gains, since the voice guard changes \emph{how} arguments are restored (Appendix~\ref{sec:qualitative}) rather than guaranteeing the largest score delta.

Error-category counts (mean per segment by MQM category, Appendix Table~\ref{tab:errorcats}) are dominated by mistranslation and omission in every condition; the full swarm shows a modest omission reduction relative to \texttt{no\_ellipsis}/\texttt{baseline}, alongside a small rise in flagged mistranslation spans, since the critique loop rewrites text and not every rewrite is judged an improvement. The tail of the per-verse score distribution (Appendix Figure~\ref{fig:tail}) confirms the system's effect is concentrated in the low tail (P10, minimum, \%$<80$ in Table~\ref{tab:ablation}) rather than a uniform shift of the whole distribution, since the mean is compressed near 85--86 for every condition.

\subsection{Judge Validation: Self-Preference and Score-Arithmetic Reliability} \label{sec:judge-validation}
Table~\ref{tab:judgeval} summarizes the evaluation-methodology findings motivating the two-judge split (full per-condition self-preference in Appendix Table~\ref{tab:judgeval-full}). \textbf{Self-preference is large and universal}: on every one of the 12 conditions, $J_{\text{route}}$ scores the output $1.9$–$6.8$ MQM higher than $J_{\text{report}}$. Reporting $J_{\text{route}}$'s own scores, as a single-judge design would, would inflate every headline number by an amount that itself varies by condition, biasing cross-condition comparison; to our knowledge, this is one of the more direct in-pipeline measurements of LLM-judge self-preference reported for a low-resource MT system. \textbf{The judge's own arithmetic disagrees with its stated rubric on $96.4\%$ of scored items} (mean $|\Delta|=16.37$ between the self-reported score and the score recomputed from its own error spans), which is why we report the recomputed score as the metric and both scores as a diagnostic. Recomputing MQM is a deliberate departure from the published GEMBA-MQM protocol, so the two are not numerically interchangeable with prior work using the standard prompt. Test-retest reliability has a within-item SD of $1.45$ MQM, smaller than the smallest reportable effect in Table~\ref{tab:contrasts}; the reporting judge's failure rate is $0.41\%$.

\textbf{Few-shot prompting reduces, but does not eliminate, this arithmetic gap.} Motivated by the concern that a zero-shot judge may simply be miscalibrated, we re-score a stratified subsample, the same 100 items per condition, paired across all twelve, with a three-shot variant of the judge prompt. This reduces the arithmetic-disagreement rate from $96.4\%$ to $79.3\%$ (mean $|\Delta|$: $16.37\to8.46$) but leaves most items still disagreeing, reinforcing that recomputing the score from error spans is not an optional refinement. Zero-shot and few-shot scores on the same items correlate only moderately (Pearson $r=0.47$, Spearman $\rho=0.60$), indicating the judge's output is itself non-trivially prompt-sensitive. At $n=100$ per condition the pre-registered contrasts do not reach significance here ($n\approx146$ is needed to detect a 2-point effect), so we treat this as a calibration diagnostic, not a second significance test.

\begin{table}[h]
\centering
\small
\caption{Judge validation summary under the zero-shot judge prompt (the few-shot comparison above is a separate, $n=100$-per-condition subsample). Self-preference range is $J_{\text{route}}-J_{\text{report}}$ over the 12 conditions.}
\begin{tabular}{lr}
\toprule
Judge diagnostic & Value \\
\midrule
Reporting judge & \texttt{openai:gpt-4o-mini} \\
Judge failure rate & 0.41\% \\
Arithmetic disagrees with rubric & 96.4\% (mean $|\Delta|$ = 16.37) \\
Test--retest SD (repeated calls) & 1.45 \\
Self-preference range (min--max, 12 conditions) & +1.94 to +6.79 \\
\bottomrule
\end{tabular}
\label{tab:judgeval}
\end{table}

\subsection{Tool Utility} \label{sec:tool-utility}
Table~\ref{tab:toolutil} reports usage for the \texttt{TRANSLATION\_MEMORY} tool: 5,513 queries, $4.9\%$ hit rate, zero transport errors. Its measured effect ($+0.01$ MQM, Table~\ref{tab:contrasts}) is below the noise floor, representing a genuine negative result rather than an artifact: fuzzy translation-memory lookup rarely finds a close precedent for an unseen verse, and when it does, the 26B-parameter generalist model already appears to carry comparable lexical knowledge parametrically.

\begin{table}[h]
\centering
\small
\caption{Tool utility, aggregated over all conditions in which the tool inventory was active.}
\input{paper/tables/tool_utility.tex}
\label{tab:toolutil}
\end{table}

\section{Conclusion}

We introduced \texttt{MITRA-Swarm}, a multi-agent architecture injecting explicit linguistic structure (Sandhi splitting, register-aware discourse ellipsis restoration, and a two-judge quality architecture) into an otherwise monolithic foundation-model pipeline for classical Sanskrit. On a pre-registered, stratified $N=1{,}000$ benchmark, the swarm produces small but statistically robust gains over a monolithic baseline and a call-count-matched self-refinement baseline, with the isolated discourse-restoration ablation as the single largest component effect. Equally central is what the benchmark says about evaluation itself: LLM-judge self-preference is large and universal (2--7 MQM, every condition), and a preflight decontamination sweep is directly falsifiable and, here, clean. A foundation-model system for a morphologically rich, low-resource classical language should be evaluated with rigorous self-scrutiny; we release the pipeline, matrix, and diagnostics so that findings can be independently reproduced. Future work: gold linguistic annotation to measure each agent against the phenomenon it targets rather than only aggregate MQM; extension to Classical Tibetan and Pāli; and a multi-judge reporting panel to bound the single-judge design's residual uncertainty.

\section*{Limitations}

\begin{itemize}
    \item \textbf{Effects are small relative to the metric's range.} Every condition scores 85.2--85.9 of 100; the effect concentrates in the low tail, not a uniform mean shift, and should not be extrapolated to a metric with more headroom.
    \item \textbf{A single reporting judge is a single point of failure.} Despite the validation in Section~\ref{sec:judge-validation}, the headline metric rests on one model; a multi-judge panel (future work) would strengthen this.
    \item \textbf{The noise floor bounds generation variance, not judge variance}, since \texttt{baseline}/\texttt{baseline\_dup} differ only in seed and are judged at temperature 0 with caching.
    \item \textbf{Recomputing MQM from error spans departs from the published GEMBA-MQM protocol} (96.4\% disagreement with the model's own arithmetic, Section~\ref{sec:judge-validation}), so it is not numerically comparable to prior work reporting the raw score.
    \item \textbf{\texttt{bo\_n} decodes at temperature 0.7 vs.\ 0.0 for every other condition}: this is inherent to best-of-$N$, but a genuine confound.
    \item \textbf{Single-prompt reasoning baselines}: While \texttt{self\_refine} and \texttt{bo\_n} provide compute-matched controls, exploring unified single-prompt Chain-of-Thought reasoning (prompting a single frontier model to perform Sandhi analysis and discourse tracking in explicit scratchpad tokens before emitting a translation) represents a valuable direction for future baseline comparisons.
    \item \textbf{Ensemble aggregation vs.\ routing}: The parallel drafting array selects the top candidate via $J_{\text{route}}$; evaluating multi-candidate synthesis/aggregation models rather than tournament routing remains an open architectural alternative.
\end{itemize}

\paragraph{Ethical considerations.} Multi-agent inference raises compute cost per query relative to single-pass models (Table~\ref{tab:compute}), acceptable for offline research but warranting caching and hardware-efficiency care before deployment; translating culturally and religiously loaded terminology (\textit{Dharma}, \textit{Śūnya}) risks interpretive drift, which the voice guard and independent evaluation aim to mitigate. We credit and respect the licenses of the open assets used: \texttt{MitraSamgraha} (CC-BY-SA 4.0; \citealp{nehrdich2026mitrasamgrahacomprehensiveclassicalsanskrit}) and the \texttt{Gemma} model family (Gemma Open License; \citealp{gemma4google2026}).


\clearpage
\bibliographystyle{acl_natbib}
\bibliography{custom}

\clearpage
\appendix

\section{Qualitative Analysis: Register-Aware Voice Preservation} \label{sec:qualitative}
Table~\ref{tab:qualitative_case} shows an unedited pipeline output pair for a \textit{legal} (Manusmṛti, voice-preserving) verse. Both conditions produce fluent, accurate English; the full swarm's Ellipsis Restorer, informed that this work is voice-preserving, rewrites the target clause as an impersonal passive ("the mind should be turned away...") rather than an invented active third-person subject ("one should turn the mind away..."), matching the impersonal register rather than converting it to a personal directive. We report this as a mechanism demonstration, not a large per-item MQM swing: per-item routing-judge scores are a diagnostic signal, not a reported metric (Section~\ref{sec:two-judge}); the statistically assessed effect of this mechanism is the register-stratified pattern (Section~\ref{sec:results}) and the isolated ablation (Section~\ref{sec:results}, trend 2), not any single verse.

\begin{table}[h]
\centering
\small
\caption{Register-aware voice preservation, an unedited output pair (\textit{Manusmṛti} 2.98, item \texttt{GSD36manu\_test:130}).}
\begin{tabular}{lp{9.5cm}}
\toprule
\textbf{Source} & \textit{indriyārtheṣu sarveṣu na prasajyeta kāmataḥ / atiprasaktiṃ caiteṣāṃ manasā saṃnivartayet //} \\ \midrule
\textbf{Gold Reference} & ``...let him carefully obviate an excessive attachment to them, by (reflecting on their worthlessness in) his heart.'' \\ \midrule
\textbf{\texttt{no\_ellipsis} (guard inactive)} & ``...furthermore, \underline{one should turn} the mind away from excessive attachment to these.'' \\ \midrule
\textbf{\texttt{full\_swarm} (\textit{legal} $\to$ passive)} & ``...furthermore, \underline{the mind should be turned} away from excessive attachment to these.'' \\ \bottomrule
\end{tabular}
\label{tab:qualitative_case}
\end{table}

\section{Supplementary Tables}
\label{sec:appendix-tables}

\begin{table}[h]
\centering
\scriptsize
\caption{Ablation matrix: active components per configuration. Guide = Sandhi Parser; Ellip. = Ellipsis Restorer; Crit. = critique-refinement loop; Tools = tool-selector inventory; TM = translation-memory prefetch; QA = automated QA audit pass. Referenced from Section~\ref{sec:ablation_configs}.}
\resizebox{\linewidth}{!}{\input{paper/tables/config_matrix.tex}}
\label{tab:configmatrix}
\end{table}

\begin{table}[h]
\centering
\caption{Full 12-condition register breakdown, MQM by register and configuration ($J_{\text{report}}$); the condensed discussion in Section~\ref{sec:results} is derived from this.}
\resizebox{\linewidth}{!}{\input{paper/tables/register_breakdown.tex}}
\label{tab:registers-full}
\end{table}

\begin{table}[h]
\centering
\small
\caption{Full per-verse token budget and latency by configuration (measured); Table~\ref{tab:compute} in the main text shows the three key conditions. Referenced from Section~4.4.}
\input{paper/tables/compute_budget.tex}
\label{tab:compute-full}
\end{table}

\begin{table}[h]
\centering
\scriptsize
\caption{Errors per segment by MQM category, $J_{\text{report}}$. Referenced from Section~\ref{sec:results}.}
\input{paper/tables/error_categories.tex}
\label{tab:errorcats}
\end{table}

\begin{table}[h]
\centering
\scriptsize
\caption{Full per-condition judge-validation diagnostics; self-preference is $J_{\text{route}}-J_{\text{report}}$. Table~\ref{tab:judgeval} in the main text summarizes this. Referenced from Section~\ref{sec:judge-validation}.}
\input{paper/tables/judge_validation.tex}
\label{tab:judgeval-full}
\end{table}

\begin{figure}[h]
\centering
\includegraphics[width=0.6\linewidth]{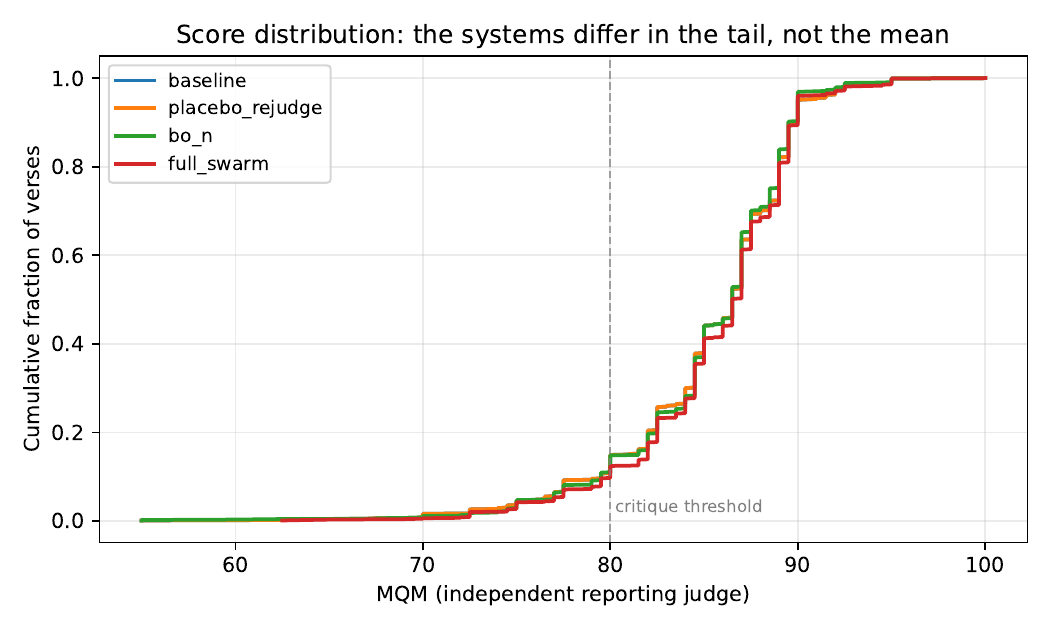}
\caption{Tail of the per-verse MQM distribution, full swarm vs.\ monolithic baseline. Referenced from Section~\ref{sec:results}.}
\label{fig:tail}
\end{figure}

\section{Agent System Prompts and Prompt Templates}
\label{sec:appendix}

In this appendix, we provide the exact system instructions, prompt templates, and JSON schemas used by the various subagents in \texttt{MITRA-Swarm} to ensure full reproducibility of the agentic behaviors.

\subsection{Morphological Pre-processing (Sandhi Parser Agent)}
The Sandhi Parser Agent is responsible for decomposing complex phonetic mergers and nominal compounds before translation begins. Its output is threaded into every downstream translator call via a typed \texttt{TranslationRequest.guide} field (Section~3.2).
\begin{verbatim}
System Instruction:
You are an expert Sanskrit grammarian.
Analyze the input Sanskrit text. Identify
complex compound words (Samasa) and
phonetic joins (Sandhi). Write down the
compound word splits and explain the
individual components briefly to assist a
translator. Format your output as a concise
split analysis. If there are no complex
compounds, reply: 'No complex compounds.'

Prompt Template:
Analyze compounds/sandhi in this text:

{source_chunk}
\end{verbatim}

\subsection{Discourse Argument Restoration (Ellipsis Restorer Agent)}
This agent operates over a sliding window to interpolate omitted arguments (zero-anaphora) into the draft translation. In production, the \texttt{genre} field passed below is \texttt{"legal"} whenever the source work ID is flagged \texttt{voice\_preserving} in the register table (covering the \textit{legal} and \textit{sutra} registers), and the literal register name otherwise (Section~\ref{sec:registers}).
\begin{verbatim}
System Instruction:
You are an expert classical language editor.
Sanskrit frequently omits pronouns,
subjects, or verbs (ellipses) when they are
clear from context. Review the source
Sanskrit text, the surrounding discourse
context, and the draft English translation.
If the source genre is impersonal or
injunctive (e.g. legal/sutra), prefer
impersonal or passive restorations over
inventing an active third-person subject.
Identify if any implied arguments are
missing in the draft English translation. If
so, write down the restored/improved
translation. If the draft translation is
already complete, accurate, and fluent,
return the draft translation EXACTLY as is.
Return only the final English text without
explanations or headers.

Prompt Template:
Source Sanskrit:
{source}

Source genre: {genre}

Draft English Translation:
{draft_translation}

Surrounding Discourse Context:
  - Preceding Chunk: {prev_chunk}
  - Succeeding Chunk: {next_chunk}

Output the restored translation:
\end{verbatim}

\subsection{Quality Evaluation Judge (GEMBA-MQM)}
Both $J_{\text{route}}$ and $J_{\text{report}}$ (Section~\ref{sec:two-judge}) use this prompt; the reported metric is the deterministic \texttt{score} recomputed from \texttt{errors} using the fixed weights below, not the model's own top-line number, which is retained separately as \texttt{model\_reported\_score} for the judge-validation diagnostic in Section~\ref{sec:judge-validation}.
\begin{verbatim}
System Instruction:
You are an expert translation quality evaluator
trained in MQM (Multidimensional Quality
Metrics). You identify translation errors from
classical languages into English and assign
severity levels. Be rigorous but fair. Respond
strictly in the JSON schema requested.

Prompt Template:
Source language: {lang_name}
Source text:
"""
{source}
"""

Candidate translation (English):
"""
{translation}
"""

Identify every translation error. For each,
choose:
  - category: one of [accuracy/mistranslation,
    accuracy/omission, accuracy/addition,
    accuracy/untranslated, fluency/grammar,
    fluency/register, fluency/spelling,
    fluency/punctuation, style/awkward,
    terminology, other]
  - severity: one of [minor, major, critical]
  - span: the offending English substring
    (verbatim)
  - explanation: one short sentence

Then compute an overall MQM score. Use the standard weights:
  critical = -25, major = -5, minor = -1
  score = max(0, 100 + sum(weights))

Return ONLY valid JSON matching:
{
  "errors": [
    {"category": "...", "severity": "...", "span": "...",
     "explanation": "..."}
  ],
  "score": <number 0-100>,
  "rationale": "<one sentence summary>"
}
\end{verbatim}

\subsection{Critique and Tool-Selector Agents}
When a translation falls below the quality threshold, the Critique Agent diagnoses the errors, and the Tool Selector Agent retrieves necessary context strictly from the tools reported as available in the current environment (Section~\ref{sec:tools}); for the benchmark in this paper, that inventory reduces to \texttt{TRANSLATION\_MEMORY}.
\begin{verbatim}
Critique Agent System Instruction:
You are a senior translation editor. A Sanskrit-to-English translation
has received a low evaluation score from our automated judge. Perform a
detailed critique. Identify errors, mistranslations, omissions, or
terminology mismatches in the draft compared to the Sanskrit source.
Suggest specific corrections.

Tool Selector Agent System Instruction:
You are a research coordinator routing translation tasks to tools. Review
the Sanskrit source text and the critique. Determine which tools from
the inventory below are needed to resolve the issues. Only tools listed
here are available in this environment; do not request any other tool.
Available Tools (environment-reported; varies by run):
1. TRANSLATION_MEMORY: fuzzy-match the source against a decontaminated
   parallel corpus for sentence-level precedents (query = Sanskrit text).
2. DICTIONARY: look up a headword in the offline lexicon snapshot, if
   one is configured for this run (query = single word in IAST).
3. ALIGNMENT: Tibetan/Classical Chinese parallel alignment, if an
   alignment index is configured for this run (query = Sanskrit keyword).

You must select between 0 and 3 tool calls. Return your selection ONLY
as a JSON list of dictionaries, e.g.:
[
  {"tool": "TRANSLATION_MEMORY", "query": "citraśravaḥ tama havante"}
]
Do not output markdown code blocks or any
explanation.
\end{verbatim}

\newpage

\input{checklist.tex}
\end{document}

%% file: paper/tables/ablation_main.tex
\begin{tabular}{lrrrrrrr}
\toprule
Configuration & $N$ & MQM & SD & P10 & P25 & Min & \%$<$80 \\
\midrule
\texttt{baseline} & 1000 & 85.27 & 5.67 & 80.0 & 84.0 & 35 & 9.0 \\
\texttt{baseline\_dup} & 1000 & 85.27 & 5.53 & 80.0 & 84.0 & 35 & 8.5 \\
\texttt{placebo\_rejudge} & 1000 & 85.26 & 5.71 & 80.0 & 84.0 & 35 & 8.9 \\
\texttt{bo\_n} & 1000 & 85.58 & 5.51 & 80.0 & 84.0 & 40 & 8.7 \\
\texttt{self\_refine} & 1000 & 85.39 & 5.56 & 80.0 & 84.0 & 35 & 8.3 \\
\texttt{guide\_only} & 1000 & 85.23 & 5.65 & 80.0 & 84.0 & 40 & 8.9 \\
\texttt{guide\_ellipsis} & 999 & 85.91 & 5.03 & 80.0 & 84.0 & 60 & 7.4 \\
\texttt{no\_guide} & 999 & 85.91 & 5.20 & 80.0 & 84.0 & 60 & 7.1 \\
\texttt{no\_ellipsis} & 1000 & 85.47 & 5.22 & 80.0 & 84.0 & 55 & 8.3 \\
\texttt{no\_critique} & 999 & 85.91 & 5.03 & 80.0 & 84.0 & 60 & 7.4 \\
\texttt{no\_tools} & 999 & 85.89 & 5.29 & 80.0 & 84.0 & 40 & 7.6 \\
\texttt{full\_swarm} & 999 & 85.90 & 5.29 & 80.0 & 84.0 & 40 & 7.6 \\
\bottomrule
\end{tabular}

%% file: paper/tables/contrasts.tex
\begin{tabular}{lrrlrrr}
\toprule
Contrast & $N$ & $\Delta$MQM & 95\% CI & $p_{\text{adj}}$ & $d_z$ & W/L/T \\
\midrule
Full swarm vs monolithic baseline & 999 & +0.63$^{*}$ & [+0.30, +0.97] & 0.002 & +0.12 & 397/294/308 \\
Re-judging alone (placebo) & 1000 & -0.01$^{\dagger}$ & [-0.07, +0.01] & 1.000 & -0.02 & 2/2/996 \\
Full swarm vs best-of-N (compute matched) & 999 & +0.32 & [-0.05, +0.67] & 0.460 & +0.06 & 363/333/303 \\
Full swarm vs self-refine (specialisation) & 999 & +0.51$^{*}$ & [+0.16, +0.85] & 0.029 & +0.09 & 389/301/309 \\
Morphosemantic guide & 999 & -0.01$^{\dagger}$ & [-0.27, +0.23] & 1.000 & -0.00 & 170/179/650 \\
Ellipsis restoration & 999 & +0.44$^{*}$ & [+0.12, +0.73] & 0.029 & +0.09 & 323/232/444 \\
Tool retrieval & 999 & +0.01$^{\dagger}$ & [+0.00, +0.03] & 1.000 & +0.04 & 2/0/997 \\
Critique loop (isolated) & 999 & -0.02$^{\dagger}$ & [-0.14, +0.09] & 1.000 & -0.01 & 24/27/948 \\
Baseline -> guide & 1000 & -0.04$^{\dagger}$ & [-0.38, +0.27] & 1.000 & -0.01 & 252/256/492 \\
Guide -> guide + ellipsis & 999 & +0.69$^{*}$ & [+0.39, +1.00] & 0.001 & +0.14 & 324/221/454 \\
\bottomrule
\end{tabular}

\footnotesize $^{*}$ significant after Holm--Bonferroni over 10 pre-registered contrasts. $^{\dagger}$ below the measured noise floor of 0.21 MQM (the difference between two identical pipelines under different decoding seeds) and therefore not distinguishable from run-to-run variation.

%% file: paper/tables/tool_utility.tex
\begin{tabular}{lrrrr}
\toprule
Tool & Calls & Hit rate & Errors & Latency (ms) \\
\midrule
\texttt{TRANSLATION\_MEMORY} & 5513 & 0.05 & 0 & 513 \\
\bottomrule
\end{tabular}

%% file: paper/tables/config_matrix.tex
\begin{tabular}{lccccccl}
\toprule
Configuration & Guide & Ellip. & Crit. & Tools & TM & QA & Notes \\
\midrule
\texttt{baseline} & -- & -- & -- & -- & -- & -- & -- \\
\texttt{baseline\_dup} & -- & -- & -- & -- & -- & -- & seed $+1$ \\
\texttt{placebo\_rejudge} & -- & -- & -- & -- & -- & -- & re-judge only \\
\texttt{bo\_n} & -- & -- & -- & -- & -- & -- & $N=5$ \\
\texttt{self\_refine} & -- & -- & \checkmark & -- & -- & \checkmark & unspecialised \\
\texttt{guide\_only} & \checkmark & -- & -- & -- & -- & -- & -- \\
\texttt{guide\_ellipsis} & \checkmark & \checkmark & -- & -- & -- & -- & -- \\
\texttt{no\_guide} & -- & \checkmark & \checkmark & \checkmark & \checkmark & \checkmark & -- \\
\texttt{no\_ellipsis} & \checkmark & -- & \checkmark & \checkmark & \checkmark & \checkmark & -- \\
\texttt{no\_critique} & \checkmark & \checkmark & -- & -- & \checkmark & \checkmark & -- \\
\texttt{no\_tools} & \checkmark & \checkmark & \checkmark & -- & \checkmark & \checkmark & -- \\
\texttt{full\_swarm} & \checkmark & \checkmark & \checkmark & \checkmark & \checkmark & \checkmark & -- \\
\bottomrule
\end{tabular}

%% file: paper/tables/register_breakdown.tex
\begin{tabular}{lrrrrrrrrrrrr}
\toprule
Register & \texttt{baseline} & \texttt{baseline\_dup} & \texttt{bo\_n} & \texttt{full\_swarm} & \texttt{guide\_ellipsis} & \texttt{guide\_only} & \texttt{no\_critique} & \texttt{no\_ellipsis} & \texttt{no\_guide} & \texttt{no\_tools} & \texttt{placebo\_rejudge} & \texttt{self\_refine} \\
\midrule
epic & 86.0 & 85.8 & 85.8 & 86.0 & 85.8 & 85.9 & 85.8 & 86.0 & 86.4 & 86.0 & 86.0 & 85.9 \\
kavya & 85.4 & 85.5 & 85.3 & 85.8 & 85.7 & 85.4 & 85.7 & 85.4 & 85.7 & 85.8 & 85.4 & 85.4 \\
legal & 85.1 & 84.7 & 85.0 & 85.4 & 85.1 & 85.0 & 85.1 & 84.8 & 85.2 & 85.4 & 85.1 & 84.6 \\
sutra & 85.7 & 85.8 & 85.9 & 86.0 & 86.0 & 85.5 & 86.0 & 85.8 & 86.2 & 86.0 & 85.7 & 86.1 \\
vedic & 83.6 & 83.9 & 83.7 & 84.5 & 84.6 & 83.4 & 84.7 & 83.5 & 84.4 & 84.6 & 83.6 & 83.7 \\
\bottomrule
\end{tabular}

%% file: paper/tables/compute_budget.tex
\begin{tabular}{lrrrr}
\toprule
Configuration & Prompt tok. & Compl. tok. & Total/verse & Latency (s) \\
\midrule
\texttt{baseline} & 91 & 49 & 140 & 2.5 \\
\texttt{baseline\_dup} & 91 & 48 & 140 & 2.4 \\
\texttt{bo\_n} & 455 & 258 & 713 & 12.8 \\
\texttt{full\_swarm} & 828 & 462 & 1290 & 7.1 \\
\texttt{guide\_ellipsis} & 738 & 430 & 1169 & 5.1 \\
\texttt{guide\_only} & 482 & 389 & 871 & 2.5 \\
\texttt{no\_critique} & 738 & 430 & 1169 & 5.1 \\
\texttt{no\_ellipsis} & 726 & 472 & 1198 & 7.3 \\
\texttt{no\_guide} & 415 & 114 & 529 & 6.8 \\
\texttt{no\_tools} & 797 & 458 & 1255 & 6.8 \\
\texttt{placebo\_rejudge} & 91 & 49 & 140 & 2.5 \\
\texttt{self\_refine} & 164 & 89 & 253 & 4.9 \\
\bottomrule
\end{tabular}

%% file: paper/tables/error_categories.tex
\begin{tabular}{lrrrrrr}
\toprule
Configuration & addition & mistranslation & omission & untranslated & awkward & grammar \\
\midrule
\texttt{baseline} & 0.07 & 1.62 & 1.35 & 0.02 & 0.02 & 0.13 \\
\texttt{baseline\_dup} & 0.07 & 1.62 & 1.36 & 0.02 & 0.02 & 0.13 \\
\texttt{bo\_n} & 0.07 & 1.63 & 1.37 & 0.01 & 0.02 & 0.13 \\
\texttt{full\_swarm} & 0.06 & 1.68 & 1.31 & 0.01 & 0.02 & 0.13 \\
\texttt{guide\_ellipsis} & 0.06 & 1.69 & 1.31 & 0.01 & 0.02 & 0.13 \\
\texttt{guide\_only} & 0.06 & 1.66 & 1.32 & 0.02 & 0.02 & 0.15 \\
\texttt{no\_critique} & 0.07 & 1.69 & 1.31 & 0.01 & 0.02 & 0.13 \\
\texttt{no\_ellipsis} & 0.06 & 1.70 & 1.33 & 0.01 & 0.02 & 0.14 \\
\texttt{no\_guide} & 0.07 & 1.66 & 1.34 & 0.01 & 0.02 & 0.11 \\
\texttt{no\_tools} & 0.06 & 1.68 & 1.31 & 0.01 & 0.02 & 0.13 \\
\texttt{placebo\_rejudge} & 0.07 & 1.62 & 1.35 & 0.02 & 0.02 & 0.13 \\
\texttt{self\_refine} & 0.07 & 1.68 & 1.35 & 0.01 & 0.02 & 0.12 \\
\bottomrule
\end{tabular}

\footnotesize Errors per segment, by MQM category, as judged by the independent reporting judge.

%% file: paper/tables/judge_validation.tex
\begin{tabular}{lr}
\toprule
Judge diagnostic & Value \\
\midrule
Reporting judge & \texttt{openai:gpt-4o-mini} \\
Judge failure rate & 0.41\% \\
Judge arithmetic disagrees with rubric & 96.4\% (mean $|\Delta|$ = 16.37) \\
Test--retest SD (same input, repeated calls) & 1.45 \\
Self-preference, \texttt{baseline} ($J_{\text{route}} - J_{\text{report}}$) & +2.84 \\
Self-preference, \texttt{baseline\_dup} ($J_{\text{route}} - J_{\text{report}}$) & +2.72 \\
Self-preference, \texttt{bo\_n} ($J_{\text{route}} - J_{\text{report}}$) & +6.79 \\
Self-preference, \texttt{full\_swarm} ($J_{\text{route}} - J_{\text{report}}$) & +5.43 \\
Self-preference, \texttt{guide\_ellipsis} ($J_{\text{route}} - J_{\text{report}}$) & +4.49 \\
Self-preference, \texttt{guide\_only} ($J_{\text{route}} - J_{\text{report}}$) & +1.94 \\
Self-preference, \texttt{no\_critique} ($J_{\text{route}} - J_{\text{report}}$) & +4.48 \\
Self-preference, \texttt{no\_ellipsis} ($J_{\text{route}} - J_{\text{report}}$) & +5.12 \\
Self-preference, \texttt{no\_guide} ($J_{\text{route}} - J_{\text{report}}$) & +5.77 \\
Self-preference, \texttt{no\_tools} ($J_{\text{route}} - J_{\text{report}}$) & +5.46 \\
Self-preference, \texttt{placebo\_rejudge} ($J_{\text{route}} - J_{\text{report}}$) & +2.90 \\
Self-preference, \texttt{self\_refine} ($J_{\text{route}} - J_{\text{report}}$) & +5.65 \\
\bottomrule
\end{tabular}